\documentclass[11pt,letterpaper,twocolumn]{article}
\usepackage[letterpaper,margin=.75in,columnsep=.25in]{geometry}
\usepackage[T1]{fontenc}
\usepackage{times,amsmath,amssymb,newtxmath,bm}
\usepackage{graphicx,booktabs,tabularx,microtype,cite,xcolor}
\definecolor{linkblue}{HTML}{285981}
\usepackage[breaklinks,colorlinks,allcolors=linkblue]{hyperref}
\usepackage[font=small,labelfont=bf]{caption}
\makeatletter
\renewcommand\section{\@startsection{section}{1}{0pt}
  {-2.4ex plus -.5ex minus -.2ex}{1.0ex plus .2ex}
  {\normalfont\large\bfseries}}
\renewcommand\subsection{\@startsection{subsection}{2}{0pt}
  {-1.8ex plus -.4ex minus -.2ex}{.65ex plus .2ex}
  {\normalfont\normalsize\bfseries}}
\makeatother

\newcommand{\bb}{\mathcal B}
\title{Can LLMs Recover Visual Evidence\\Lost to Token Compression?}
\hypersetup{pdftitle={Can LLMs Recover Visual Evidence Lost to Token Compression?}}
\author{%
  \normalsize Zhiyao Lin\textsuperscript{1}\quad
  Emily Parker\textsuperscript{2}\quad
  Yucheng Shen\textsuperscript{1}\quad
  Gabriel Adineera\textsuperscript{2}\quad
  Ethan Bennett\textsuperscript{2}\\[4pt]
  \small \textsuperscript{1}Boston College\qquad
  \textsuperscript{2}Texas A\&M University-Commerce
}
\hypersetup{pdfauthor={Zhiyao Lin, Emily Parker, Yucheng Shen, Gabriel Adineera, Ethan Bennett}}

\makeatletter
\def\@maketitle{%
  \newpage
  \null
  \vskip .5em%
  \begin{center}%
    \let\footnote\thanks
    {\LARGE \@title\par}%
    \vskip 1em%
    {\large
      \lineskip .5em%
      \begin{tabular}[t]{c}%
        \@author
      \end{tabular}\par}%
  \end{center}%
  \par\vskip .75em%
}
\makeatother
\date{}
\let\originalthebibliography\thebibliography
\renewcommand{\thebibliography}[1]{%
  \originalthebibliography{#1}%
  \setlength{\itemsep}{2pt plus .2pt minus .2pt}%
}
\begin{document}
\setlength{\jot}{3pt}
\setlength{\abovedisplayskip}{7pt plus 2pt minus 2pt}
\setlength{\belowdisplayskip}{7pt plus 2pt minus 2pt}
\setlength{\abovedisplayshortskip}{4pt plus 1pt}
\setlength{\belowdisplayshortskip}{5pt plus 1pt minus 1pt}
\maketitle
\begin{abstract}
Visual token compression can remove small details or make surviving evidence difficult for a language decoder to use. Recent weight-space programming results motivate asking whether an LLM can repair such failures by editing a compressed model. We formulate this question at a fixed token count and distinguish edits before the visual bottleneck from edits downstream of an unchanged representation. For identical compressed inputs and the same question, downstream editing cannot give different answers to two images that the bottleneck has made indistinguishable. Upstream editing can change this equivalence relation while preserving token count. We specify a restricted weight-editing protocol, paired held-out evaluation, and controls for edit location, magnitude, and increased visual capacity. A reanalysis of published Adaptive-VoCo results motivates these controls: its normalized advantage over a reproduced one-token baseline is strongly influenced by ScienceQA-Image and uses a variable one-to-four-token budget. These results characterize the starting compression problem. The effectiveness of the proposed LLM-guided repairs remains an empirical question requiring frozen candidate edits and new held-out evaluations.
\end{abstract}
\section{Introduction}
A compressed image representation may retain the shape of a building while losing the small address needed to answer a question about it. It may also retain useful distinctions that the language decoder fails to use. These failures suggest different repair strategies. Increasing the number of tokens can expose more evidence, whereas changing the mapping into a fixed number of tokens can change which distinctions survive.

Modern visual instruction models connect patch encoders to language models through representations that can be long and costly~\cite{llava,llava15,llavanext,qwen25}. Learned bottlenecks, including VoCo-LLaMA and hybrid token compression, expose a compact visual prefix to text generation~\cite{voco,hybrid}. Adaptive-VoCo selects the prefix length from image statistics~\cite{adaptive}. Such designs make a resource decision and a representation decision at once. A successful fixed-budget repair must improve the representation or its use while holding the resource decision constant.

Weight-space programming offers a way to investigate this possibility. Takehi et al.\ describe LLMs that inspect and directly modify neural parameters~\cite{wsp_sparks}. Zhang et al.\ apply this interface to robot policies and report localized repair of lexical instruction grounding, with paired controls and explicit regressions~\cite{zhang2026can}. In vision--language models, repairs of deliberately planted faults can depend on numerical traces left by fault construction~\cite{wsp_vlmrepair}. These findings motivate a carefully delimited question about naturally occurring compression errors: can an LLM choose useful parameter edits when the visual token count stays fixed?

We develop a formulation and an evaluation protocol for this question. First, we identify a boundary imposed by identical compressed representations and explain why editing before the bottleneck can alter it. Second, we specify a calibration-only editor that writes a restricted parameter patch and freezes its decision before held-out evaluation. Third, we define paired measures of recovery and regression together with controls that isolate edit direction, location, and token allocation. Published compression scores provide context for the required comparisons. Detailed score tables, proofs, and allocation analyses appear in the supplement.

Our account establishes what a recovery claim would need to measure. The supplied record contains no LLM-produced parameter patch or corresponding recovery evaluation. The proposed protocol can be applied to a released checkpoint without interpreting a new title or a stronger compression baseline as evidence of a successful weight edit.
\section{Related Work}
\paragraph{Compression and variable capacity.}
BLIP-2 uses a query transformer to connect visual and language representations~\cite{blip2}. LLaVA-Mini, TokenPacker, and nested query models explore compact visual prefixes with different mechanisms for preserving information~\cite{llavamini,tokenpacker,mqt,matryoshka}. Pruning methods such as FastV, SparseVLM, and VisionZip remove or consolidate visual context~\cite{fastv,sparsevlm,visionzip}. ATP-LLaVA, VisionThink, and RUTA make input-dependent decisions about visual resources~\cite{atp,visionthink,ruta}. The complete taxonomy in the supplement distinguishes latent bottlenecks, patch selection, and adaptive rate decisions. A patch to the compressor must be evaluated at the same count and insertion point as its unedited counterpart.

\paragraph{Direct weight editing.}
The robotics study of Zhang et al.~\cite{zhang2026can} is a direct methodological precedent. Its successful edits are localized to token embeddings and their effects are measured on matched held-out episodes. Temporal Weight Surgery studies diagnosed errors in video grounding and compares learned edits to size-matched random changes~\cite{wsp_temporal}. Recursive Self-Editing applies repeated LLM-authored updates to the editor model itself~\cite{wsp_selfwrite}. The VLM repair study of Aluguvelly~\cite{wsp_vlmrepair} exposes a different issue: planted parameter corruption can leave cues that naturally occurring failure need not provide. Compression errors arise under an intact trained checkpoint, so our protocol starts from the released model and records the numerical format of every edited tensor.

\paragraph{Evaluating preserved evidence.}
MM-CoT examines visual reasoning separately from answer correctness~\cite{mmcot}.
VTC-Bench, UniPruneBench, and PAA-Bench show why task accuracy, retained-token count, and speed need careful joint interpretation~\cite{vtc,uniprune,paa}. Their evaluation perspective motivates testing recoveries on paired examples and measuring realized cost. Counterfactual VLM fine-tuning also motivates examining dependence on visual evidence~\cite{cfvlm}. We further hold the image-level budget assignment fixed when comparing weights, so a better routing policy cannot explain an apparent representation repair.
\begin{figure*}[t]
\centering
\includegraphics[width=\textwidth]{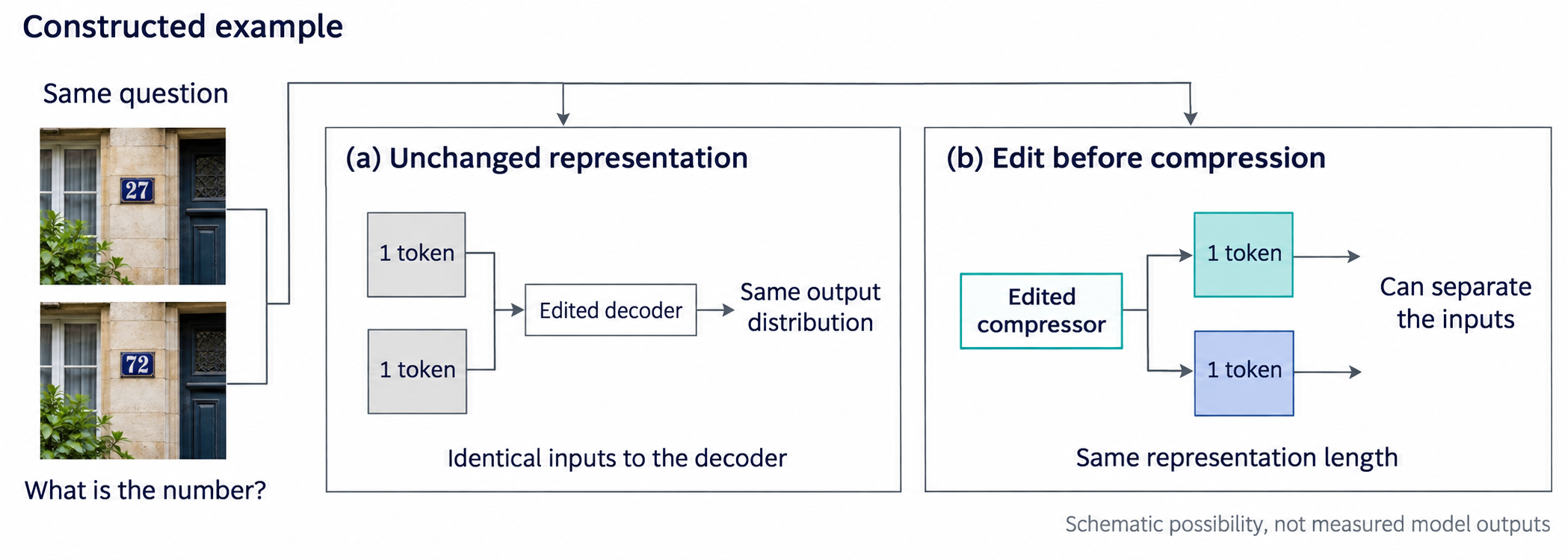}
\caption{Constructed illustration of the intervention boundary. If two images produce identical complete representations for the same question, a shared downstream decoder edit gives the same output distribution. Editing the upstream map can instead change the representations at the same token count. Different colored tiles denote distinct latent values. The images illustrate the mathematical assumption and do not record model predictions.}
\label{fig:boundary}
\end{figure*}

\section{What Can a Fixed-Budget Edit Recover?}
\label{sec:boundary}
Let $E_\phi$ encode an image and $C_{\theta,K}$ map its features to $K$ latent tokens. A language decoder $D_\psi$ produces an answer,
\begin{equation}
Z_{\theta,K}(I)=C_{\theta,K}(E_\phi(I)),\qquad
\hat y=D_\psi(Z_{\theta,K}(I),q).
\label{eq:model}
\end{equation}
The compressed representation includes every image-dependent quantity available to the decoder. If an implementation also exposes image-dependent positions, masks, or cached states, they belong in $Z$. Fixed token count constrains representation length and leaves the values of its entries free to change.

\subsection{A boundary for downstream editing}
Consider two images $I_1,I_2$ with the same question $q$ and different correct answers. If
\[
Z_{\theta,K}(I_1)=Z_{\theta,K}(I_2),
\]
a deterministic decoder with any shared downstream edit must return the same answer for both. The same statement applies to its output distribution under input-independent randomness. One member of a balanced pair must then remain incorrect. This conclusion follows because the edited decoder receives identical inputs. It concerns exact collisions in the accessible representation. Small Euclidean distance or high feature similarity alone does not establish a collision or an error bound.

Figure~\ref{fig:boundary} illustrates the distinction. An edit to $\theta$ acts earlier. It changes $Z_{\theta,K}$ even when $K$ is fixed and can separate previously identical representations. This is a possibility allowed by the interface, rather than a guarantee about a particular model. A useful empirical distinction is therefore whether an edit changes the preserved visual distinctions, changes the decoder's use of them, or merely changes answer priors. Supplement~\ref{supp:proofs} gives the finite-set statement and a task-relevance example.

\subsection{The target population}
For evaluation example $i$, let $u_i,c_i,e_i\in\{0,1\}$ be correctness for the full-visual reference, unedited compressed model, and edited model. All three use the same image, question, evaluator, and decoding specification. Define
\begin{equation}
\mathcal G=\{i:u_i=1,\ c_i=0\},\qquad
\mathcal P=\{i:c_i=1\}.
\label{eq:sets}
\end{equation}
The set $\mathcal G$ identifies observed failures relative to the full-visual model. Membership does not prove irreversible information loss because the models may use different masks or parameterizations. It provides a reproducible operational target. The primary endpoint remains performance on the whole prespecified test set, with $\mathcal G$ and $\mathcal P$ used to explain individual changes. Ground-truth test outcomes never guide edit selection.

\subsection{A finite-set accuracy ceiling}
For a fixed question, partition a finite image set $\mathcal I$ into classes $\mathcal C$ with identical decoder-accessible representations. If the reference answer is $y(I,q)$, the best deterministic downstream decoder has accuracy at most
\begin{equation}
\frac{1}{|\mathcal I|}\sum_{\mathcal C}
\max_y\bigl|\{I\in\mathcal C:y(I,q)=y\}\bigr|.
\label{eq:ceiling}
\end{equation}
The ceiling follows by choosing one answer per class. A randomized decoder is a mixture of such choices and obeys the same expected bound. Nonuniform example weights replace the counts by probability mass. This result states the relevant assumption explicitly: the entire accessible representation, rather than a selected hidden feature, must coincide.

A two-coordinate construction illustrates the distinction. Suppose the question requests $x_2$ from an input $(x_1,x_2)\in\{0,1\}^2$ and the one-scalar compressor is $z_\theta=\theta_1x_1+\theta_2x_2$. With $\theta=(1,0)$, inputs $(0,0)$ and $(0,1)$ collide and any downstream decoder shares its answer. The upstream choice $\theta=(0,1)$ preserves the requested coordinate with one scalar. The representation length is identical. This construction demonstrates an available mechanism for task-specific recovery, while learning the change in an actual VLM remains an empirical challenge.

A count of continuous tokens is therefore different from a discrete information budget. Dtype, precision, noise, and architecture matter for a quantitative capacity argument. A measured improvement should be tied to the evidence requested by the question and tested for regressions on other questions. The construction also predicts that preserving $x_2$ may harm questions about $x_1$. It does not imply a single small edit can recover every distinction under aggressive compression.
\begin{figure*}[t]
\centering
\includegraphics[width=\textwidth]{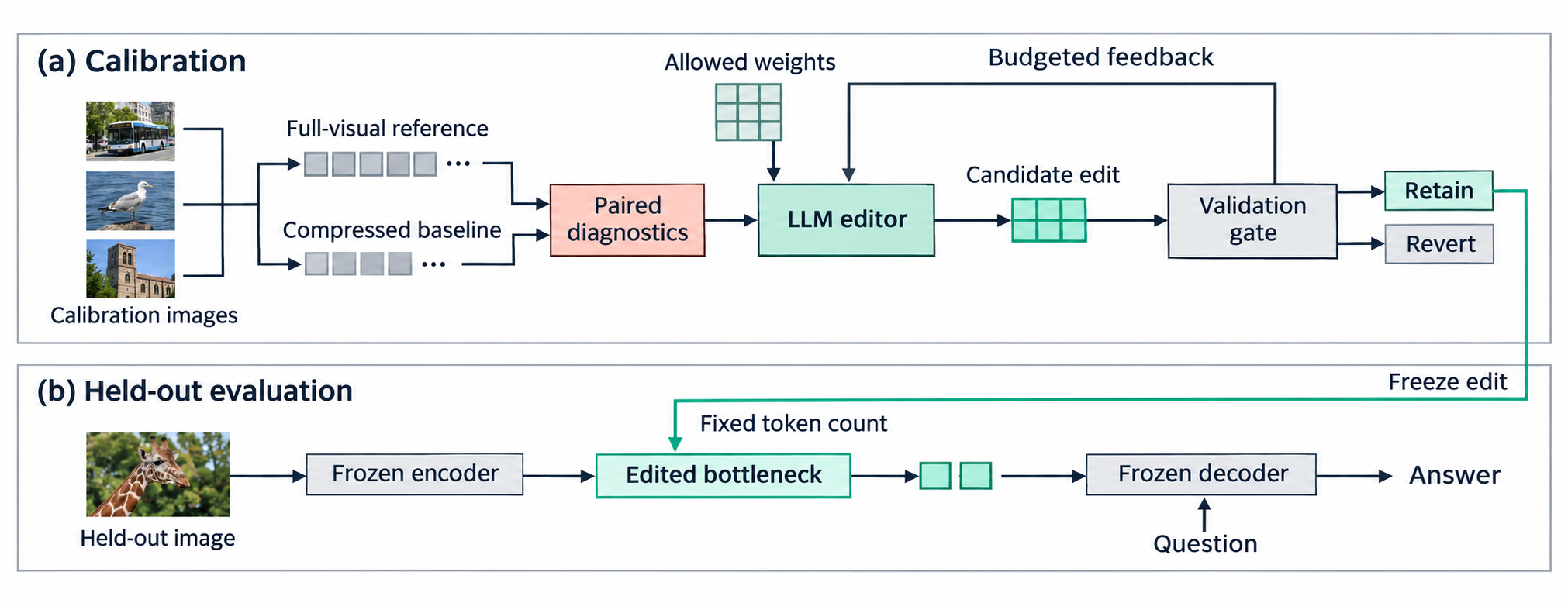}
\caption{Proposed fixed-budget repair protocol. Calibration diagnostics guide an LLM that writes a restricted parameter patch. The gate uses calibration and validation data to retain a candidate or restore the original weights. Held-out evaluation starts after this decision is frozen. The images and parameter grids illustrate the interface.}
\label{fig:repair}
\end{figure*}

\section{A Restricted LLM Repair Protocol}
\label{sec:repair}
\subsection{Intervention and available evidence}
We propose editing a declared subset of compressor or visual-projector tensors, leaving the encoder, language decoder, and token-count policy fixed in the primary comparison. The exact module depends on the chosen checkpoint. For a compressor implemented inside the language model, an experiment must identify the affected parameters and account for their use at other positions. A broad shared-layer edit cannot be interpreted as a compressor-only intervention.

The editor receives tensor names, shapes, dtypes, the allowed locations, and paired calibration diagnostics. Useful diagnostics include answers and task scores at full and compressed settings, the selected counts, and a declared set of activations. Original images may be included if the editor can inspect them. That access, any descriptions of the desired evidence, and all reference answers must be recorded. A separate validation split selects among candidates, and a test split is unavailable throughout the editing session. Images shared across questions remain in one split.

\subsection{Candidate construction and selection}
An editor session reads weights and writes a candidate patch in a persistent CPU workspace. Model execution goes through a logged forward evaluator with a declared allowance. The permitted operations include static tensor inspection, projections, or row/column updates. The evaluated protocol excludes gradient fitting and training loops. It still uses calibration observations and forward feedback, both of which contribute to cost.

For an allowed index set $\mathcal W$, a patch $\Delta$ satisfies
\begin{equation}
\operatorname{supp}(\Delta)\subseteq\mathcal W,\qquad
\operatorname{shape}(\theta+\Delta)=\operatorname{shape}(\theta).
\label{eq:patch}
\end{equation}
The evaluator checks finite values, dtype and shape compatibility, the edited parameter count, and prespecified magnitude limits before running a candidate. Limits are relative to the original tensors and are chosen on development data. Store original hashes and reload original weights before each independent candidate to avoid accumulating accidental changes.

Validation uses a fixed selection objective,
\begin{equation}
J_{\rm val}(\Delta)=
\frac{1}{n_{\rm val}}\sum_i(e_i-c_i)
-\lambda_{\rm reg}\frac{1}{|\mathcal P_{\rm val}|}
 \sum_{i\in\mathcal P_{\rm val}}(1-e_i).
\label{eq:gate}
\end{equation}
The regression term is omitted when $\mathcal P_{\rm val}$ is empty. The coefficient and a retention rule are declared before test evaluation. A stricter rule can require a minimum validation gain with an explicit regression ceiling. The session returns either a frozen patch or a revert decision. Reporting all sessions preserves the denominator of attempted repairs.

\subsection{Deployment semantics}
Freeze a candidate's tensor values before testing. Each held-out example is encoded again using the edited compressor at its original count $K_i$. Edited results are paired with the unedited output at that same $K_i$. If a policy originally assigns one, two, or four tokens, cache those hard assignments and reuse them in both arms. This prevents a changed representation from silently changing the route.

A downstream-only edit is an optional second arm with its own whitelist. It tests how much useful information remains accessible in unchanged compressed features. An edit that re-encodes an image has a different intervention point from an edit to a cached representation, and the runtime record should say which operation was rerun. The final patch adds no new module when it replaces values in existing tensors, but latency and memory still require measurement.
\section{Evidence and Evaluation}
\label{sec:eval}
\subsection{Published compression evidence}
The existing Adaptive-VoCo record~\cite{adaptive} contains seven benchmark score rows and aggregate timing curves. It reports an adaptive one-to-four-token model against a reproduced one-token VoCo-LLaMA baseline. The normalized mean is $89.3\%$ versus $81.0\%$. Recalculation from the displayed task scores gives the sensitivity in Table~\ref{tab:context}. ScienceQA-Image accounts for about $65\%$ of the normalized advantage against the reproduced baseline. The published one-token VoCo-LLaMA row~\cite{voco} yields a different comparison.

\begin{table}[t]
\centering\small
\begin{tabular*}{\columnwidth}{@{\extracolsep{\fill}}lrr@{}}
\toprule
One-token comparison & All tasks & Excl.\ SQA-I\\
\midrule
Reproduced VoCo-LLaMA & $+8.32$ & $+3.38$\\
Published VoCo-LLaMA & $+5.64$ & $-2.32$\\
\bottomrule
\end{tabular*}
\caption{Adaptive minus one-token mean retention (percentage points), recalculated from published compression scores. These differences characterize the baseline evidence and contain no weight-editing measurements.}
\label{tab:context}
\end{table}

The native scores improve on six of seven benchmarks against the reproduced baseline and decrease slightly on POPE. The adaptive and fixed runs also use different supported token counts. This record motivates measuring both capacity and task-specific changes. It cannot estimate the outcome of an LLM-authored weight patch, and the public source archive provides no such patch. Complete original task values and the latency-curve reanalysis are in Supplement~\ref{supp:reanalysis}.

\subsection{Paired recoveries and regressions}
For a frozen candidate, let $g$ count compressed failures that become successes and $\ell$ count successes that become failures on $n$ test examples. Report
\begin{equation}
\Delta_{\rm all}=\frac{g-\ell}{n},\qquad
R_{\mathcal G}=\frac{\sum_{i\in\mathcal G}e_i}{|\mathcal G|},
\label{eq:recovery}
\end{equation}
and the regression rate on $\mathcal P$. If either subset is empty, its conditional rate is undefined and the report should give the count. Separate changes within $\mathcal G$ from changes on examples the full-visual reference also misses. Show task-level scores alongside any normalized aggregate. For metrics such as corpus-level F1, run the original evaluator on the paired prediction files.

For binary outcomes on independent units, an exact paired McNemar comparison uses the $g+\ell$ discordant pairs. If an image produces several questions, resampling whole images respects that dependence. Sessions evaluated on one common test set do not create independent test populations. Report uncertainty over examples and variation across independent editor sessions separately. Prespecify the tested family of endpoints and account for multiple comparisons when making significance claims.

\begin{table}[t]
\centering\small
\begin{tabularx}{\columnwidth}{@{}lX@{}}
\toprule
Comparison arm & Intervention\\
\midrule
Original & Original weights and recorded $K_i$\\
Bottleneck edit & Change upstream map at the same $K_i$\\
Decoder edit & Keep $Z$ fixed and change downstream use\\
Random direction & Match edited support and magnitude\\
Higher-token run & Increase the available visual capacity\\
\bottomrule
\end{tabularx}
\caption{Intervention arms for a prospective repair evaluation. The higher-token arm supplies a capacity reference. The primary weight comparisons keep each example's original count.}
\label{tab:repair_controls}
\end{table}

\subsection{Controls needed for attribution}
Table~\ref{tab:repair_controls} summarizes the interventions. The primary control is the original checkpoint at exactly the same per-image token count. Match the editor's forward allowance and candidate-selection rule when comparing search procedures. Random-direction patches match the edited locations and magnitude, while random-location patches probe the importance of the location chosen. Each control also receives a held-out evaluation, including its regressions. Copying an editor's chosen patch norm alone does not match the full search process if the editor has tried many candidates.

A higher-token run gives a capacity reference. Re-encoding with a larger image or giving the decoder access to original patches should also be labeled as increased-evidence conditions. They help bound practical recovery but change the primary budget. A decoder-only patch tests downstream use of the original compressed representation. These arms distinguish where a change operates without requiring that all errors have a single cause. The supplement specifies a common comparison manifest.

Synthetic fault tests require an additional check. Parameter rescaling or injected noise can leave numerical signatures detectable by an editor~\cite{wsp_vlmrepair}. Naturally occurring compression failures are the primary target here. If injected faults are used as a diagnostic, retain matched dtype, precision, and rounding controls and report them separately from intact-checkpoint results.

\subsection{Resource measurements}
Record editor-model calls, forward examples evaluated, CPU/GPU time, and the number and size of rejected candidates. Measure inference under a shared request order, hardware, precision, generation limit, and batching policy. Report encoder, compressor, prefill, and generation costs when measurable, plus end-to-end cost. The number of latent tokens determines part of the workload, while attention kernels, KV-cache management, and generated length affect realized cost~\cite{transformer,flash,paged,computeoptimal}. Supplement~\ref{supp:allocation} retains the allocation control for experiments that deliberately allow token counts to vary.

\subsection{Crossed intervention analysis}
An optional mechanism test crosses a frozen bottleneck patch with a separately frozen decoder patch. Evaluate the original model, each single patch, and their combination at the same recorded counts. Let $A_{uv}$ denote accuracy with the bottleneck edit enabled when $u=1$ and the decoder edit enabled when $v=1$. Their score interaction is
\begin{equation}
I_{\rm edit}=A_{11}-A_{10}-A_{01}+A_{00}.
\label{eq:interaction}
\end{equation}
A positive interaction means that the combined gain exceeds the sum of the individual gains on that population. Negative values indicate subadditivity, which may include ceiling effects. This comparison tests compatibility at the score level. Inspect the paired predictions to determine which errors contribute, because an interaction alone does not locate the missing visual information. Freeze both patches and the combination before held-out evaluation and report all four arms, including a combination that fails to improve.

\subsection{Interpreting the possible outcomes}
The intervention arms support several distinguishable conclusions. If a bottleneck patch repairs held-out errors at recorded $K_i$ while a matched decoder-only arm cannot, the result is consistent with a useful upstream change. It still requires checking that the bottleneck patch has not modified shared decoder behavior or opened an additional visual path. If both arms improve, the unedited representation may already carry useful evidence that its original decoder underuses. Similar gains by matched random patches would weaken attribution to the editor's chosen direction.

A successful repair on $\mathcal G$ with many regressions on $\mathcal P$ can have negative net value. This is why the all-example endpoint precedes conditional recovery. Changes on examples missed by the full-visual reference can reflect improvements in answer priors as well as visual grounding. A controlled visual contrast, template holdout, and stable general-ability set help examine those alternatives. Reporting only the most successful session would also conceal the reliability of the editing procedure.

Finally, failure to find a useful patch under one allowed tensor set and search allowance identifies a limit of that evaluated procedure. It does not prove that the compressed model contains no recoverable evidence. The whitelist, feedback access, precision, and editor model must accompany the outcome so that another run can test a different intervention without silently changing the question.

\paragraph{Reversibility and transfer.}
A zero-patch run should reproduce the original deterministic predictions. After each candidate, restore the original tensors and repeat a fixed calibration batch. A discrepancy indicates that the comparison may include changed caches, preprocessing, numerical precision, or other state. Preserve both the replacement tensors and the restoration hashes so that a third party can replay the intervention.

Evaluate a prespecified range of edit magnitudes on the same parameter support. Set the range and any retention rule using validation data, then freeze them before testing. Include the zero-patch reference and report paired repairs and regressions at each magnitude. This reveals whether the apparent recovery is stable or depends on a narrow scaling choice.

Transfer should be evaluated along the axis implicated by the proposed repair. A patch targeting small text can be tested on held-out strings and layouts, while an object-detail patch can be tested on new instances and question templates. Include questions about unaffected regions of the same images to detect collateral changes. Report transfer failures alongside successes and retain the original count for each example. This makes a narrowly useful patch distinguishable from an improvement that depends on a familiar vocabulary or presentation.
\section{Discussion and Conclusion}
The formulation separates two meanings of recovery. Editing a compressor can preserve task evidence that its original mapping discarded at the same token count. Editing downstream weights can improve the use of a fixed representation, subject to the exact-collision boundary. Neither outcome can be inferred from aggregate accuracy alone. Image pairs with the same question and a controlled visual difference offer one way to examine dependence on evidence rather than answer priors.

Text reading and fine detail motivate held-out tests on TextVQA and OCRBench, alongside the existing multimodal tasks~\cite{textvqa,ocrbench}. Question variants and counterfactual images can reveal whether an apparent repair changes grounding~\cite{cfvlm,mmcot}. Inference-time refinement is another relevant condition. Top-Down Semantic Refinement studies planned image-caption refinement~\cite{req_tdsr}, while Self-Rewarded Multimodal Coherent Reasoning targets coherence and visual grounding in reasoning~\cite{req_selfreward}. Record any such inference procedure as part of the decoding condition. Calibration-selected improvements can still overfit a vocabulary, image style, or question template. Hold out images, templates, and appropriate task groups, and retain a general-ability set to expose regressions. Results should identify the supported checkpoint, compression mechanism, token count, and numerical format.

LLM-guided compression repair is a testable direction suggested by recent direct weight editing~\cite{zhang2026can,wsp_sparks}. This study supplies a constrained intervention, an information boundary, and a paired evaluation protocol. The original compression scores motivate that test and provide reproducible context. A positive answer to the title requires new frozen parameter patches that recover held-out errors at the original token count while accounting for regressions and measured cost\label{maintextend}.%
\clearpage
\begingroup
\fontsize{9}{10}\selectfont
\raggedright

\endgroup
\clearpage
\appendix
\setcounter{page}{1}
\renewcommand{\thepage}{S\arabic{page}}
\setcounter{figure}{0}
\renewcommand{\thefigure}{S\arabic{figure}}
\setcounter{table}{0}
\renewcommand{\thetable}{S\arabic{table}}
\setcounter{equation}{0}
\renewcommand{\theequation}{S\arabic{equation}}
\renewcommand{\theHequation}{supp.\arabic{equation}}
\renewcommand{\theHfigure}{supp.\arabic{figure}}
\renewcommand{\theHtable}{supp.\arabic{table}}
\section*{Supplementary Material}
\label{supp:start}
\fontsize{10}{12}\selectfont
\noindent Supplement to \emph{Can LLMs Recover Visual Evidence Lost to Token Compression?} Citation numbers refer to the bibliography in the accompanying main paper.
\par
\section{Information Boundary and Pairing}
\label{supp:proofs}
\subsection{The finite-set collision statement}
Fix a question $q$ and let $\mathcal I$ be a finite set of images with reference answers $y(I,q)$. Partition the images into equivalence classes according to equality of the complete decoder-accessible representation $Z_{\theta,K}(I)$. A deterministic downstream decoder returns one answer per class. Its maximum accuracy on a uniformly weighted set is therefore
\begin{equation}
\operatorname{Acc}_{\max}
=\frac{1}{|\mathcal I|}
\sum_{\mathcal C}\max_y
\bigl|\{I\in\mathcal C:y(I,q)=y\}\bigr|.
\end{equation}
Any mixture of deterministic predictions also satisfies this bound in expectation. For nonuniform weights, replace the counts by probability mass. The statement allows a shared downstream parameter edit of arbitrary size. It assumes that the decoder has no additional image-dependent input or persistent state distinguishing members of a class.

\subsection{Task relevance at fixed representation length}
Consider a two-dimensional input with components $(x_1,x_2)$, and one latent scalar $z_\theta=\theta_1x_1+\theta_2x_2$. Suppose all questions ask for the binary value $x_2$. Setting $\theta=(1,0)$ discards the answer-relevant coordinate. A shared downstream edit cannot distinguish $(0,0)$ from $(0,1)$ because both have latent value zero. Replacing $\theta$ with $(0,1)$ retains the relevant value at the same latent length. This construction shows how upstream editing can repair a task-specific bottleneck. It supplies no algorithm for finding the edit in a trained VLM and gives no guarantee about transfer to questions that require $x_1$.

\subsection{Paired counts and uncertainty}
Let $n_{ab}$ count examples with original correctness $a$ and edited correctness $b$. Then $g=n_{01}$, $\ell=n_{10}$, and
\[
\Delta_{\rm all}=(n_{01}-n_{10})/n .
\]
Conditional on $g+\ell$ under the usual paired null, the number favoring the edit is binomial with success probability $1/2$. Its exact two-sided tail probability gives a paired McNemar comparison for independent binary units. A confidence interval for the population effect requires an appropriate interval or resampling construction and need not coincide with a permutation interval over shuffled budgets.

When several questions share an image, bootstrap image IDs and retain all associated questions in each draw. Aggregate accuracy may weight questions equally, while resampling clusters by image. Report that distinction and the number of independent images. If several editor sessions share one held-out population, their gains are paired across sessions as well. Report each session's count table and a paired across-session comparison when needed. Successive candidates chosen by the same feedback loop are an adaptive search history and should not be counted as independent sessions.

\subsection{Budget matching during weight repair}
If an original router chooses $K_i$, a primary parameter-repair comparison evaluates both original and edited weights at that same recorded $K_i$. Matching only the mean count can conceal shifts between easy and difficult examples. Preserving the full histogram still permits reassignment, which is why the primary repair arm retains the per-example counts. A separate routing experiment can use the shuffled control in Supplement~\ref{supp:allocation}. This separates changing the representation at a chosen rate from choosing a more useful rate.
\section{Editor Interface and Evaluation Manifest}
\label{supp:implementation}
\subsection{Checkpoint and data contract}
A concrete run needs the checkpoint identifier, original tensor hashes, tokenizer and image preprocessing, target compression mechanism, supported token counts, and a whitelist of editable parameter locations. Record head aggregation, spatial indexing, and the layer at which text loses access to original patch tokens. A nominal one-token bottleneck can still carry image information through additional caches or masks. Include those channels in the representation definition and instrument them when making a downstream-collision claim.

Partition by image ID before constructing calibration failures. A calibration split may expose examples on which full-visual and compressed runs disagree. A separate validation set selects the patch, and held-out images are used after the decision is frozen. Make any template, object category, OCR vocabulary, or dataset-level holdout explicit. Extra descriptions written after observing a test failure can leak the missing evidence into the editor prompt, even when the tensor patch is small.

The editor manifest should include its exact model/version, prompts, tool schema, state inherited from earlier sessions, decoding settings, and all forward feedback. Count both the number of requests and the number of model examples or generated tokens inside a request. Larger batches of feedback change the information available to the editor. If a calibration reference is a model prediction rather than a gold answer, record its source and allow the corresponding uncertainty.

\subsection{Patch construction and validation}
Before each candidate, restore original values and verify their hashes. Apply the candidate in a disposable model instance and check allowed tensor names, dtype, shape, finite values, edited entries, Frobenius norms, row/column support, and any prespecified relative norm bound. A patch accepted by these checks can still harm capability. The validation objective and retention rule determine whether to keep it.

The policy must identify whether the editor may read original image pixels, full-visual activations, compressed activations, gradients, answer text, or correctness labels. The proposed primary protocol permits logged calibration observations and static tensor operations with a budgeted forward evaluator. It excludes gradient fitting and iterative training. An implementation must enforce that distinction at its execution interface. Labeling an arbitrary numerical search as reasoning alone would omit part of the method. If a starting router uses a Gumbel relaxation~\cite{gumbel,concrete}, log its hard deployment choices as well as the expected counts used during training. Adaptive training strategies can affect the starting checkpoint~\cite{matagent}, so comparisons should identify its full training provenance.

Freeze the selected patch and its selection record before any held-out call. Verify that the target's inference program and token-count assignments match the original run. Candidate count, wall time, evaluator use, and rejected or reverted sessions belong in the report. Retain/revert selection should not remove unsuccessful sessions from the denominator.

\subsection{Comparison arms}
\begin{table*}[t]
\centering\small
\begin{tabularx}{\textwidth}{@{}lXXX@{}}
\toprule
Arm & Weights & Per-image count & What it tests\\
\midrule
Original compressed & Original checkpoint & Recorded $K_i$ & Starting accuracy and cost\\
LLM-edited bottleneck & Selected allowed patch & Same $K_i$ & Fixed-budget representation repair\\
Random direction & Matched support and norm & Same $K_i$ & Direction conditional on edit location\\
Random location & Matched size and norm & Same $K_i$ & Sensitivity to chosen parameter support\\
Decoder-only edit & Separate downstream whitelist & Same $K_i$ & Use of unchanged compressed features\\
Higher-token reference & Original checkpoint & Explicitly increased & Additional visual capacity\\
Full-visual reference & Declared full-visual setup & Original full prefix & Observed reference task performance\\
\bottomrule
\end{tabularx}
\caption{Proposed experimental arms and their interpretation. Each arm needs its own held-out prediction files and cost record. Random controls can be evaluated as individual patches or as a matched search procedure, with the choice reported.}
\label{tab:arms}
\end{table*}
A random direction can be generated within the same tensor support and rescaled to match the candidate's norm. Multiple random draws probe sensitivity to direction. If the LLM has selected among many candidates using validation, an equally budgeted random-search control better compares search procedures. Report the best-of-search rule, the complete search allowance, and test the selected control once. Conditioning a control on support chosen by the editor answers a narrower question than matching the full selection process.

A decoder-only arm must keep the visual representation unchanged, including image-dependent masks, caches, and positions. A higher-resolution or higher-token arm provides a practical reference but changes the available visual evidence and compute. A reference with shared weights is preferable for isolating rate effects. Independently trained fixed-rate checkpoints measure a different combination of training and inference choices.

\subsection{Visual and numerical diagnostics}
A recovery claim benefits from examples that retain the same scene and question while changing a small answer-relevant feature. Record both original and edited predictions, the reference answer, and the exact pixels changed. Such examples can probe grounding but their construction must not be given to the editor after selecting test cases. Multi-view evidence aggregation also appears in Tri-MARF~\cite{trimarf}. DepthSSC uses depth-spatial alignment and voxel adaptation for monocular semantic scene completion~\cite{req_depthssc}. A 3D repair extension would require its own geometry-aware evaluation and an explicit account of the views available.

Record the storage precision before and after editing. Where synthetic corruption is used, calculate and round in the original dtype as a separate control. A high-precision construction can expose identifying traces even if it has a small effect on activations~\cite{wsp_vlmrepair}. Natural compression errors avoid that construction but can still leak dataset identity or answer priors through diagnostic prompts.

\subsection{Task and resource scope}
The NTIRE 2026 challenge report evaluates image-space $\times4$ super-resolution~\cite{req_ntire}. If restored images are supplied to a VLM as an auxiliary condition, record that preprocessing and its cost. It changes the image input, whereas the primary weight intervention reuses the prescribed images. The shared-assistance setting in Markets of Minds~\cite{markets} motivates recording the opportunity cost of a long editing search.

MM-CoT motivates checking visual reasoning traces separately from final correctness~\cite{mmcot}. LLaVA-OneVision provides another possible checkpoint family~\cite{onevision}. Their presence in the literature does not establish compatibility with a particular weight patch. Porting the intervention requires a fresh tensor whitelist and matched baseline.

Agent resource allocation offers a related systems perspective. KABB chooses expert assistance~\cite{kabb}. Yin et al. review the broader VLM architectures and capabilities to which collaboration can be applied~\cite{vlmsurvey}. GAM-Agent uses uncertainty-guided visual collaboration~\cite{gam}. OSC further studies knowledge alignment during communication among collaborating agents~\cite{req_osc}. The resource consumed by a weight editor includes its own model calls and calibration evaluations. Inference savings must be reported together with that editing cost and the number of future requests over which it would be amortized.
\section{Complete Published Compression Evidence}
\label{supp:reanalysis}
\label{sec:evidence}
\begin{table*}[t]
\centering
\small
\setlength{\tabcolsep}{5pt}
\begin{tabular*}{\textwidth}{@{\extracolsep{\fill}}lrrrrrrrrr@{}}
\toprule
Configuration & Tokens & GQA & VQAv2 & MMBench & MME & POPE & SEED & SQA-I & Ret. (\%) \\
\midrule
Full-visual reference & 576 & 61.1 & 77.7 & 64.0 & 1487.2 & 85.0 & 57.9 & 66.5 & 100.0 \\
Q-Former~\cite{blip2,voco} & 1 & 51.1 & 63.4 & 51.7 & 1079.7 & 77.3 & 47.2 & 62.7 & 57.2 \\
Avg. pooling + linear~\cite{voco} & 1 & 52.9 & 65.0 & 55.5 & 1210.3 & 79.1 & 50.3 & 62.2 & 64.1 \\
VoCo-LLaMA (rerun)~\cite{adaptive} & 1 & 57.4 & 71.8 & 57.9 & 1241.4 & 81.5 & 48.8 & 66.3 & 81.0 \\
VoCo-LLaMA (published)~\cite{voco} & 1 & 57.0 & 72.3 & 58.8 & 1323.3 & 81.4 & 53.7 & 65.4 & 83.7 \\
\textbf{Adaptive-VoCo}~\cite{adaptive} & 1--4 & \textbf{57.6} & \textbf{72.3} & \textbf{60.7} & 1285.1 & 81.4 & 50.2 & \textbf{68.5} & 89.3 \\
\midrule
Lower reference & 1 & 37.7 & 41.2 & 22.3 & 617.3 & 53.9 & 36.9 & 60.7 & 0.0 \\
\bottomrule
\end{tabular*}
\caption{Published benchmark scores. The reproduced and externally published one-token VoCo-LLaMA rows have different sources. Retention (Ret.) is recalculated from the seven displayed scores using Eq.~\eqref{eq:retention}. SQA-I is the image subset of ScienceQA. Bold marks adaptive scores that equal or exceed every other compressed model row above the lower reference.}
\label{tab:bench}
\end{table*}
\begin{table}[t]
\centering
\small
\begin{tabular*}{\columnwidth}{@{\extracolsep{\fill}}lrrr@{}}
\toprule
Benchmark & vs.\ 1 & vs.\ 2 & vs.\ 4 \\
\midrule
GQA & $+3.5$ & $-2.0$ & $-0.7$ \\
VQAv2 & $+3.4$ & $+1.7$ & $+0.5$ \\
MMBench & $+5.6$ & $-1.2$ & $-1.7$ \\
MME & $+4.3$ & $-0.6$ & $+0.6$ \\
POPE & $-1.1$ & $-3.9$ & $-3.2$ \\
SEED & $+5.1$ & $-1.3$ & $+1.0$ \\
ScienceQA-I & $+3.7$ & $-1.6$ & $-0.2$ \\
\bottomrule
\end{tabular*}
\caption{Approximate adaptive minus fixed-rate latency (ms) by benchmark, digitized from the original timing figure~\cite{adaptive}. Positive values indicate longer adaptive runs. Submillisecond differences are descriptive.}
\label{tab:time}
\end{table}
The published Adaptive-VoCo evidence consists of seven benchmark-level scores per model, reference rows, and separate curves for fixed-rate and adaptive inference~\cite{adaptive}. No per-example predictions, rate histograms, timing traces, or trained checkpoint accompany the public TeX archive. This supplement reproduces the baseline reanalysis. None of the rows measures a weight-edited model.

\subsection{Models, benchmarks, and normalization}
The original report uses the VoCo-LLaMA setup for image encoding, language modeling, and preprocessing, and reproduces its one-token configuration as the primary comparison~\cite{adaptive,voco}. We retain that reproduced row and add the one-token row from VoCo-LLaMA's conference paper with explicit attribution. The reference, Q-Former, and average-pooling scores are carried over from the same VoCo-LLaMA comparison. Its single-token average-pooling implementation includes a linear projection. The lower reference uses a one-token bottleneck without vision-compression training, and the upper reference makes the full set of 576 visual tokens accessible. These names denote two model configurations, so individual results can exceed the upper reference score.

Counterfactual changes to an image can probe the effect of altered visual evidence~\cite{cfvlm}. The reported evaluation compares outcomes on standard benchmark splits. It covers GQA and VQAv2~\cite{gqa,vqav2}, MMBench and MME~\cite{mmbench,mme}, POPE~\cite{pope}, SEED-Bench~\cite{seed}, and the image subset of ScienceQA~\cite{scienceqa}. The benchmarks have different native scoring units. Following the original table, for benchmark $j$ we normalize model score $s_j$ using the published lower and upper reference scores $l_j,u_j$:
\begin{equation}
R_j=100\,\frac{s_j-l_j}{u_j-l_j},\qquad
\bar R=\frac{1}{7}\sum_{j=1}^{7}R_j .
\label{eq:retention}
\end{equation}
Every task has equal weight after normalization. Under this convention its raw sensitivity is proportional to $1/(u_j-l_j)$. The resulting mean is a summary of relative positions between the reference rows, while the individual task scores keep their original evaluation units. We use the reported rounded scores to compute the normalized quantities.

\subsection{Where the reported gain comes from}
Adaptive-VoCo improves GQA by $0.2$, VQAv2 by $0.5$, MMBench by $2.8$, MME by $43.7$, SEED-Bench by $1.4$, and ScienceQA-Image by $2.2$ against the reproduced one-token baseline. POPE decreases by $0.1$. The mean reference-normalized difference is $8.32$ percentage points, but the normalized task differences vary considerably. ScienceQA-Image has a reference gap of $5.8$ native score points. Its $2.2$-point improvement becomes $37.93$ points on the normalized scale and contributes $5.42$ points to the mean across all seven tasks. With ScienceQA-Image excluded, the mean advantage is $3.38$ points. The pattern warrants reporting both task scores and the normalized summary.

\begin{figure*}[t]
\centering
\includegraphics[width=\textwidth]{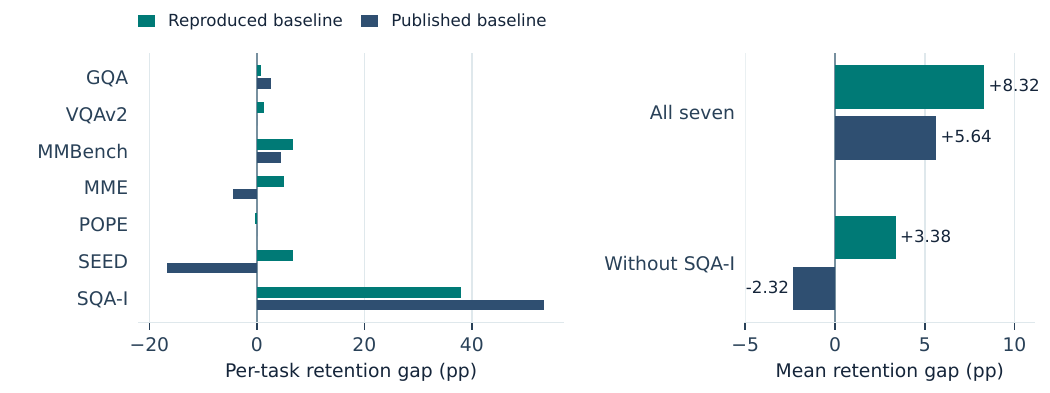}
\caption{Sensitivity of reference-normalized retention to the choice of one-token baseline and to task inclusion. Left, per-task differences between Adaptive-VoCo and each one-token row in Table~\ref{tab:bench}. Right, mean differences over all tasks and with ScienceQA-Image excluded. Both panels use the reference rows printed in the table.}
\label{fig:sensitivity}
\end{figure*}

Changing the one-token comparison also changes the pattern (Fig.~\ref{fig:sensitivity}). Against the published VoCo-LLaMA row, the adaptive row is higher on GQA, MMBench, and ScienceQA-Image, ties on VQAv2 and POPE at the printed precision, and is lower on MME and SEED-Bench. The aggregate difference is $5.64$ points with all tasks and $-2.32$ points after excluding ScienceQA-Image. These computations expose sensitivity to the baseline source. Reproduction and published results can differ in checkpoint, preprocessing, or evaluation details, and the table does not isolate those causes.

The original fixed-two and fixed-four retention curves are useful supplementary context. The adaptive curve sits above both on MMBench and ScienceQA-Image and below both on GQA, VQAv2, and MME~\cite{adaptive}. Because the realized rate mixture is unavailable, even similar benchmark totals against a single fixed-two row do not establish a matched-budget result. None of the aggregate rows reports which images became correct or incorrect when the compressed representation changed.

\subsection{Timing order across tasks}
Table~\ref{tab:time} redraws the original Fig.~2(b) timing evidence as benchmark differences using marker centers read from the published plot. These are approximate values to the nearest $0.1$~ms. The four curves cover seven benchmark means under the reported single-sample setting. The adaptive method takes longer than fixed-one on six benchmarks and less time on POPE. Its time relative to fixed-four changes direction by benchmark, and the apparent difference on ScienceQA-Image is close to the available graphical resolution. The fixed-two curve can have higher time than fixed-four. This nonmonotonicity is compatible with several causes, including the timing boundary, generated output length, or runtime variation.

A $576$-to-$K$ reduction concerns how many visual positions enter the downstream representation. It does not specify the rate of end-to-end computation or the share of memory attributable to these positions. In particular, collecting attention variance can require exposing statistics that a fused attention kernel would ordinarily leave implicit. A controlled latency claim needs measurements with that cost included. The original plot provides direction and approximate scale, but its hardware, warm-up, generation lengths, and timing variability are unavailable. We therefore report the ordering descriptively and leave uncertainty intervals to paired timing runs.
\section{Allocation Controls for Variable Rates}
\label{supp:allocation}
\label{sec:protocol}
A run that gives some images four tokens has more opportunities to preserve details than a run that always uses one. An assignment test must preserve those opportunities while changing who receives them. We describe the test for a finite validation set and make the same-model requirement explicit.

\subsection{The response matrix}
For $M$ evaluation units indexed by $i$, force each supported count $b\in\bb$ through one frozen checkpoint. Let $s_{ib}$ be the resulting unit score and $c_{ib}$ the measured resource cost, when the latter is collected. A unit may be a question or an image with several questions. When a compressed image is shared by its questions, use the image as the assignment unit. The row can then store the sum of its question scores, and the final accuracy divides by the total number of questions. Sample order, evaluator, decoding policy, and resolution remain fixed. Cache reuse across rates is allowed only for operations that are independent of the selected count. With stochastic decoding, use a prespecified rule for seeds and estimate variability across seeds separately.

Let $a_i\in\bb$ be the router's hard budget before any forced runs. Partition units into disjoint groups $G_1,\ldots,G_H$ by information chosen before looking at test outcomes. One group per benchmark gives a control for benchmark-level budget differences. If image format or resolution changes the work before compression, use groups that also match those properties where group sizes permit. A single group over all requests asks a different question, since a router that assigns more tokens to an entire benchmark can then gain from cross-benchmark allocation. Report the grouping, histograms, and number of groups with variation in $a_i$.

\subsection{An exact shuffled-budget control}
For a group $G$, write $n_G=|G|$ and $h_{Gb}=\sum_{i\in G}\mathbf{1}\{a_i=b\}$. A uniform permutation of the vector of counts within $G$ preserves the complete histogram. Consequently, it preserves both the maximum count and every statistic of the counts aggregated within that group. Each unit in $G$ receives count $b$ with probability $h_{Gb}/n_G$, even though assignments to different units are dependent. For a sum of example scores, the expected mean of the control follows by linearity:
\begin{equation}
\begin{aligned}
S_{\mathrm{route}}&=\frac{1}{M}\sum_{i=1}^{M}s_{i,a_i},\\
S_{\mathrm{shuffle}}&=\frac{1}{M}\sum_G\sum_{i\in G}
                 \sum_{b\in\bb}\frac{h_{Gb}}{n_G}s_{ib},\\
\Delta_{\mathrm{assign}}&=S_{\mathrm{route}}-S_{\mathrm{shuffle}}.
\end{aligned}
\label{eq:assign}
\end{equation}
With an image-level assignment and per-question accuracy, replace the common divisor $M$ in these expressions with the number of scored questions. The exact expectation uses the same response matrix for both arms. It needs no assumption that larger budgets always improve an answer. A negative value indicates that the learned placement is worse than the quota-preserving randomized placement on that evaluation set.

For a fixed-rate run $S_b$ of the very same checkpoint, adding and subtracting the control gives
\begin{equation}
S_{\mathrm{route}}-S_b =
\Delta_{\mathrm{assign}}+(S_{\mathrm{shuffle}}-S_b).
\label{eq:decomp}
\end{equation}
The second term changes the rate distribution without using the router's within-group information. With one group and an additive metric it is a weighted mean of the fixed-rate results minus $S_b$. In the two-rate case, let $d_i$ be the score at the higher rate minus the score at the lower one. The assignment effect is the within-group covariance of $d_i$ and the indicator that the router selected the higher rate, averaged across groups in proportion to their sizes. A router can gain only by sending larger budgets to units with larger realized gains within a group. Published fixed-rate models trained separately can help establish a practical frontier, but they cannot be substituted for the same-checkpoint outputs in Eq.~\eqref{eq:decomp}. Checkpoint differences would enter the second term.

\begin{figure*}[t]
\centering
\includegraphics[width=\textwidth]{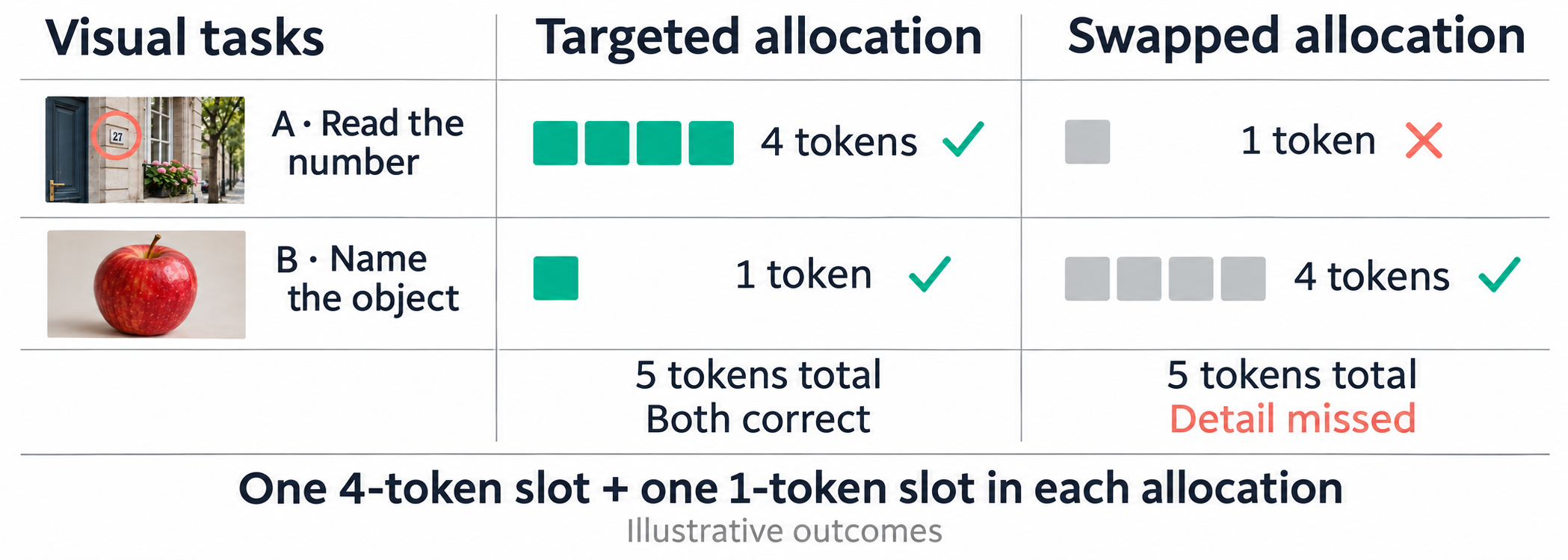}
\caption{Hypothetical allocation example. Reading a small number benefits from the larger slot, while recognizing an apple succeeds at either rate. Each allocation uses one 4-token slot and one 1-token slot. The images and outcomes illustrate the control described in the text.}
\label{fig:quota}
\end{figure*}
In Figure~\ref{fig:quota}, the two assignments answer two and one of the questions correctly. A uniform shuffle averages $75\%$, and assigning one token to both images gives $50\%$. Of the targeted assignment's gain over fixed-one, half comes from the larger slot and half from placing it on the image that benefits. These outcomes are stipulated for the example.

\paragraph{Beyond decomposable metrics.}
For scores such as corpus-level F1, a mean of row-wise F1 values generally changes the metric. Save the original predictions, permute the counts within each group, select predictions from the corresponding columns, and run the full evaluator for each permutation. Apply the same rule to metrics defined over groups of questions if the evaluator cannot be expressed as a sum over assignment units. Optional permutation quantiles describe the spread of the randomized control on the evaluation set. Population uncertainty is a separate question. It can be estimated by resampling independent images or clusters and recomputing the test together with the corresponding quota weights. The fraction of shuffles at or above the route score describes its extremeness under the randomized control. Interpreting that fraction as a significance test requires an exchangeability null conditional on the groups and quotas.

\subsection{A quota-constrained ceiling}
For an additive score, the strongest assignment with the same histogram can be found by solving
\begin{equation}
\begin{aligned}
\max_{\{x_{ib}\}}\quad
&\sum_i\sum_{b\in\bb}x_{ib}s_{ib}\\
\text{s.t.}\quad
&\sum_{b\in\bb}x_{ib}=1,\quad x_{ib}\in\{0,1\},\\
&\sum_{i\in G}x_{ib}=h_{Gb}\quad\forall\,G,b .
\end{aligned}
\label{eq:oracle}
\end{equation}
This problem has a flow formulation within each group: each example chooses one outgoing edge to a rate, and a rate accepts exactly its quota. If there are two rates, choose the largest observed gains for the larger-rate quota. With three or more rates, the largest adjacent gain depends on which alternatives remain available, so a joint assignment avoids committing the same example twice. The ceiling uses ground-truth outcomes and quantifies headroom at the observed histogram. Learning a deployable policy requires fitting only on development or training responses and evaluating on held-out units.

\subsection{From count matching to cost matching}
A preserved histogram does not guarantee the same latency. Assignments can change generation length, correlate the rate with image size, or alter the longest request in a padded batch. The exact shuffle calculation can be repeated with $c_{ib}$ in place of $s_{ib}$ for additive per-request costs and a fixed execution regime. For service throughput or tail latency, rerun the scheduled workload under each policy instead. Report the timing boundary and the actual counts for prefill and generation, and compare score at a measured cost target in addition to the controlled count comparison.

For a decoder that sees $T$ text prompt positions and $b$ compressed positions, a dense prefill layer's attention arithmetic scales with $(T+b)^2$ up to the causal masking constant. At fixed head dimensions, cached keys and values grow linearly with retained prefix length. These expressions describe language-decoder scaling and omit the cost of creating the visual features, making the routing decision, and producing the summaries. A smaller cache on the decoder side can coexist with an unchanged image encoder and substantial front-end work. The hardware, attention kernel, and batching policy matter even at the same token count~\cite{transformer,flash,paged,computeoptimal}.
\section{Extended Compression Literature}
\label{sec:related}

\paragraph{Learned visual bottlenecks.}
Patch encoders such as ViT and CLIP produce a sequence of contextualized image representations~\cite{vit,clip}. Perceiver uses latent queries to control the size of the expensive representation~\cite{perceiver}, and TokenLearner learns a small number of spatially aggregated tokens~\cite{tokenlearner}. Flamingo resamples visual features and exposes them to the language model through cross-attention~\cite{flamingo}. BLIP-2 instead uses a query transformer to bridge frozen image and language components~\cite{blip2}. These designs create new latent summaries. As a result, a latent token and a retained patch can have very different informational roles even when they occupy the same number of positions.

LLaMA-VID separates context and content representations~\cite{llamavid}. LLaVA-Mini learns to fuse visual information before presenting a small visual prefix to the language model~\cite{llavamini}. TokenPacker designs its projector to combine reduced tokens with finer local information~\cite{tokenpacker}. VoCo-LLaMA uses an attention mask and training objectives to route image evidence through learned bottleneck tokens inside the decoder~\cite{voco}. Hybrid token compression enriches the input to that bottleneck~\cite{hybrid}. Matryoshka Multimodal Models learn nested resolutions~\cite{matryoshka}, while Matryoshka Query Transformer trains interchangeable prefixes of latent queries~\cite{mqt}. Variable-width training makes a common-checkpoint comparison feasible, but leaves the choice of width to a policy or a preset. Our control measures the contribution of that choice at fixed capacity.

\paragraph{Pruning, merging, and their cost.}
DynamicViT, A-ViT, and EViT reduce computation in a vision encoder by choosing which patch representations keep participating or by reorganizing inattentive tokens~\cite{dynamicvit,avit,evit}. Token Merging groups similar tokens~\cite{tome}. In multimodal decoders, FastV uses early-layer attention to reduce the tokens processed by subsequent layers, and PyramidDrop progressively changes token retention across depth~\cite{fastv,pyramiddrop}. SparseVLM leverages text-guided visual relevance~\cite{sparsevlm}. VisionZip and LLaVA-PruMerge use importance and redundancy cues, while DivPrune and DART prioritize diversity or duplicate removal~\cite{visionzip,prumerge,divprune,dart}. Dynamic-LLaVA considers both visual and language context sparsity~\cite{dynamicllava}. These mechanisms also differ in where they pay for dense features and score candidate tokens. The final retained count is comparable as a representation length even when runtime costs differ.

\paragraph{Input-dependent rate decisions.}
Adaptive computation time allocates a variable number of recurrent steps to each input~\cite{act}. DrDiff combines dynamic expert routing and hierarchical sparse attention for long-text diffusion~\cite{req_drdiff}, providing a cross-domain example of jointly controlling interaction and expert computation. SmartTrim adapts tokens and attention heads, and ATP-LLaVA sets instance- and layer-dependent thresholds~\cite{smarttrim,atp}. Recent work has explored energy-based count selection, adaptive thresholding, and sequential selection with a stopping decision~\cite{eadaprune,adaptpruning,stepprune}. VIP-Router chooses a pruning strategy per sample~\cite{viprouter}. VisionThink starts from a lower-resolution image and can request a higher-resolution one~\cite{visionthink}. RUTA conditions token selection on the question and optimizes task utility against expected usage~\cite{ruta}. These precedents motivate evaluating the original image-only Adaptive-VoCo router against controls with identical rate frequencies. The test can also be used with a question-conditioned router when the same checkpoint supports forced-rate inference.

\paragraph{Evaluation and resource allocation.}
VTC-Bench filters compression-sensitive examples, UniPruneBench compares tasks and runtime, and PAA-Bench examines selection-sensitive accuracy and speed~\cite{vtc,uniprune,paa}. Our shuffle redistributes counts while preserving candidate representations and tests the usefulness of allocation. Prompt compression likewise examines useful content under a sequence budget~\cite{llmlingua,llmlingua2}. Expert consultation and feedback-driven workflow changes concern related resource decisions~\cite{kabb,gam,workflow}.
\end{document}